\documentclass[11pt]{article}

\usepackage[preprint]{acl}

\usepackage{times}
\usepackage{latexsym}
\usepackage[T1]{fontenc}
\usepackage[utf8]{inputenc}
\usepackage{microtype}
\usepackage{graphicx}
\usepackage{amsmath}
\usepackage{amssymb}
\usepackage{booktabs}
\usepackage{xcolor}
\usepackage{array}
\usepackage[skins,breakable]{tcolorbox}
\usetikzlibrary{decorations.pathmorphing,patterns.meta}

\newcommand{\icdsid}{SemICD}
\newcommand{\codes}{\mathcal{C} }
\definecolor{dnred}{HTML}{B03A2E}
\newcommand{\dn}[1]{\textsubscript{\textcolor{dnred}{$-$#1}}}

\definecolor{skInk}{HTML}{222222}
\definecolor{skSky}{HTML}{5AA9E6}
\definecolor{skRose}{HTML}{E8849F}
\definecolor{skPeach}{HTML}{F7A94B}
\definecolor{skMain}{HTML}{9775FA}
\definecolor{skAux}{HTML}{38B98A}

\tikzset{
  sk rough/.style={decorate,decoration={random steps,segment length=5pt,amplitude=0.45pt}},
  sk border/.style={draw=skInk,line width=0.8pt,line cap=round,line join=round,sk rough},
  sk hachure/.style={pattern={Lines[angle=50,distance=4.5pt,line width=0.5pt]},pattern color=#1!80},
}
\newcommand{\skframe}[2][6pt]{%
  \draw[sk border,rounded corners=#1] #2;
  \draw[sk border,rounded corners=#1,line width=0.5pt,
    decoration={random steps,segment length=6pt,amplitude=0.6pt}] #2;
}

\newtcolorbox{promptbox}[2][skSky]{%
  enhanced, breakable,
  frame hidden, boxrule=0pt, arc=6pt,
  colback=white, colupper=skInk,
  left=10pt, right=10pt, top=14pt, bottom=8pt,
  before skip=18pt, after skip=12pt,
  fontupper=\small,
  interior hidden,
  underlay={%
    \fill[#1!7,rounded corners=6pt]
      (frame.south west) rectangle (frame.north east);
    \skframe{(frame.south west) rectangle (frame.north east)}},
  before upper={\setlength{\parskip}{4pt}},
  overlay unbroken and first={%
    \node[anchor=west,inner xsep=6pt,inner ysep=3pt,
      font=\sffamily\bfseries\small,text opacity=0] (skT)
      at ([xshift=12pt]frame.north west) {#2};
    \fill[#1!28,rounded corners=4pt]
      (skT.south west) rectangle (skT.north east);
    \fill[sk hachure=#1,rounded corners=4pt]
      (skT.south west) rectangle (skT.north east);
    \skframe[4pt]{(skT.south west) rectangle (skT.north east)}
    \node[anchor=west,inner xsep=6pt,inner ysep=3pt,
      font=\sffamily\bfseries\small,text=skInk]
      at ([xshift=12pt]frame.north west) {#2};},
}

\newcommand{\pvar}[1]{%
  \tikz[baseline=(v.base)]{%
    \node[inner xsep=2.5pt,inner ysep=1.5pt,
      fill=skRose!22,rounded corners=2.5pt] (v)
      {\ttfamily\footnotesize\{#1\}};
    \draw[draw=skInk,line width=0.45pt,line join=round,sk rough,
      decoration={random steps,segment length=2.5pt,amplitude=0.3pt}]
      ([xshift=0.4pt]v.south west) rectangle ([xshift=-0.4pt]v.north east);
  }%
}
\newcommand{\prole}[1]{\textsf{\bfseries\color{skInk}#1:}\ }

\title{Beyond ICD Codes: Semantic IDs for Generative Medical Coding}

\author{
  \href{mailto:yihang.cheng@postgrad.otago.ac.nz}{Yihang Cheng}\thanks{Corresponding author.} \\
  \normalfont\normalsize University of Otago \\
  Dunedin, New Zealand \\
  \href{mailto:yihang.cheng@postgrad.otago.ac.nz}{\small yihang.cheng@postgrad.otago.ac.nz}
  \And
  Qian Zhang \\
  \normalfont\normalsize University of Otago \\
  Dunedin, New Zealand \\
  \href{mailto:qian.zhang@postgrad.otago.ac.nz}{\small qian.zhang@postgrad.otago.ac.nz}
  \And
  Yonghui (Andie) Huang \\
  \normalfont\normalsize Massey University \\
  Palmerston North, New Zealand \\
  \href{mailto:andie.huang@massey.ac.nz}{\small andie.huang@massey.ac.nz}
  \AND
  Baoshuo Kan \\
  \normalfont\normalsize South China Normal University \\
  Guangzhou, China \\
  \href{mailto:baoshuokan@m.scnu.edu.cn}{\small baoshuokan@m.scnu.edu.cn}
  \And
  Junyi Shen \\
  \normalfont\normalsize University of Otago \\
  Dunedin, New Zealand \\
  \href{mailto:shjy2015@gmail.com}{\small shjy2015@gmail.com}
}

\begin{document}
\maketitle

\begin{abstract}
Generative approaches to automatic International Classification of Diseases (ICD) coding use large language models (LLMs) to predict ICD codes from clinical documents. However, original ICD code strings are not ideal generation targets because the ICD hierarchy does not always reflect clinical relationships, and LLMs struggle to interpret and generate these codes. We introduce \textbf{SemICD}, a framework that trains LLMs to generate Semantic IDs (SIDs) instead. SemICD constructs SIDs by hierarchically clustering embeddings of textual code profiles, allowing semantically related codes to share prefixes while each complete SID uniquely identifies one ICD code. To align the SID tokens with clinical semantics, we employ two auxiliary tasks for bidirectional generation between textual code profiles and SIDs. Experiments on MIMIC-III and MIMIC-IV with four LLMs show that SemICD outperforms direct code generation, improving Micro-F1 by 1.34--8.00 points. Our code is available at \url{https://github.com/youwillbethebest/SemICD}.
\end{abstract}

\section{Introduction}
\label{sec:introduction}

Automatic ICD coding maps clinical documents to ICD codes used for healthcare administration, reimbursement, and analysis, aiming to reduce manual coding effort. However, accurate ICD code prediction remains challenging due to the large and long-tailed label space~\citep{mullenbach-etal-2018-explainable,huang-etal-2022-plm}. Most prior work treats ICD coding as a multi-label classification problem, assigning a score to each code~\citep{mullenbach-etal-2018-explainable,vu2020label,huang-etal-2022-plm}. More recently, generative approaches have used large language models (LLMs) to predict ICD codes from clinical documents, mainly by generating diagnosis text and mapping it to codes through retrieval over code descriptions~\citep{baksi-etal-2025-medcoder,chen2026krel} or by generating the codes directly~\citep{falis2024gpt,yuan-etal-2025-toward,wang2026posttraining}.

\begin{figure}[ht]
  \centering
  \includegraphics[width=\columnwidth]{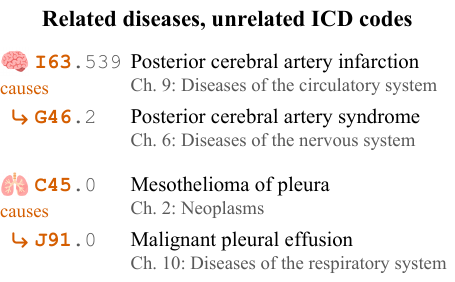}
  \caption{Two pairs of causally related diseases. The ICD-10-CM coding instructions require the two codes in each pair to be coded together, but the codes are unrelated at every level of the ICD hierarchy.}
  \label{fig:intro-related-codes}
\end{figure}

Both approaches have limitations. Mapping generated diagnosis text to codes requires a retrieval step, which can miss the correct code even when the diagnosis is correctly identified~\citep{motzfeldt-etal-2025-code,chen2026krel}. Generating ICD codes directly avoids this step but uses the original code strings as generation targets. Although ICD codes follow a hierarchical classification organized mainly by body system and etiology, this hierarchy does not always reflect clinical relationships~\citep{icd10cm2025,luo-etal-2024-corelation,lu-etal-2020-multi}. As Figure~\ref{fig:intro-related-codes} illustrates, causally related diseases that ICD-10-CM instructs coders to report together can be assigned codes from different chapters. Moreover, ICD code strings differ from the natural language on which LLMs are predominantly trained, and previous studies have found that LLMs struggle to interpret and generate these codes~\citep{soroush2024poor,ma2025mera,yin-etal-2026-icdagent,yuan-etal-2025-toward}. These limitations motivate an alternative output representation that maintains precise code identification while aligning prediction decisions with clinical semantic relationships.

Instead of generating original ICD code strings, LLMs can be trained to generate alternative identifiers that uniquely identify each code. Semantic IDs (SIDs), widely used as generation targets in generative retrieval and recommendation~\citep{tay2022transformer,rajput2023recommender}, provide such a representation by assigning items sequences of discrete tokens in which semantically related items share prefixes. This structure allows related ICD codes to share early prediction decisions while preserving their unique identities. We therefore investigate whether SIDs can serve as more effective generation targets than original ICD code strings for LLM-based ICD coding.
Specifically, we introduce \icdsid{} to train LLMs to generate SIDs for ICD coding. \icdsid{} constructs SIDs by hierarchically clustering embeddings of textual code profiles, so that semantically related codes share prefixes while each complete SID sequence identifies one ICD code. To align the introduced SID tokens with clinical semantics, we introduce two auxiliary tasks for bidirectional generation between textual code profiles and SIDs, providing explicit supervision that connects the identifiers to their clinical descriptions. At inference, constrained decoding restricts generation to valid SIDs, which are mapped back to ICD codes.

We evaluate the overall effectiveness of \icdsid{} against direct ICD code generation and examine the contributions of semantic SID assignment and auxiliary supervision. In summary, our contributions are as follows:
\begin{itemize}
\item We introduce \icdsid{}, a framework that uses SIDs instead of original code strings as generation targets for LLM-based ICD coding.
\item We construct SIDs that share prefixes across semantically related ICD codes and introduce bidirectional auxiliary tasks to align SIDs with textual code profiles.
\item Experiments on MIMIC-III and MIMIC-IV with four LLMs show consistent Micro-F1 improvements of 1.34--8.00 points. Controlled experiments demonstrate the importance of semantic assignment and the impact of SID construction.
\end{itemize}

\section{Related Work}
\label{sec:related-work}

Automatic ICD coding is commonly formulated as multi-label classification, with discriminative models predicting ICD codes from clinical notes \citep{mullenbach-etal-2018-explainable,vu2020label,huang-etal-2022-plm}. Generative approaches have also been explored for ICD coding.
Yang et al.\ generate code descriptions and map them to ICD codes \citep{yang2023multilabel}, while MedCodER combines generative extraction with retrieval and reranking \citep{baksi-etal-2025-medcoder}.
Recent work further investigates post-training for generative ICD coding \citep{wang2026posttraining}. Generative ICD coding raises the question of how ICD codes should be represented in output sequences. The choice of how to represent output affects the decomposition of code prediction into decisions and determines which of these decisions can be shared across different codes. Raw ICD strings provide concise labels, but they do not explicitly organize token sharing by clinical semantic similarity. On the other hand, code descriptions convey clinical meanings but usually require longer output sequences \citep{yang2023multilabel}. These trade-offs highlight the need for compact output representations that enable semantic sharing while maintaining the exact identity of the codes.

Semantic IDs offer a compositional representation where each item is assigned a brief sequence of discrete tokens, enabling semantically related items to share portions of their identifiers.
Previous research in generative retrieval and recommendation has developed these identifiers using methods such as clustering \citep{tay2022transformer} and residual quantization \citep{rajput2023recommender}.
Existing constructions include hierarchical clustering, as used in the Differentiable Search Index (DSI) to construct semantically structured document identifiers for generative retrieval \citep{tay2022transformer}, residual clustering with successive K-means codebooks, as used in QARM \citep{luo2024qarm}, and autoencoder residual quantization, as used in TIGER \citep{rajput2023recommender}.
Hierarchical clustering organizes sharing through common prefixes, whereas residual approaches encode items through successive quantization stages.
Beyond identifier construction, LC-Rec aligns discrete identifiers with language through bidirectional prediction between item text and identifiers \citep{zheng2024adapting}.
These differences motivate studying how SID construction affects downstream generative prediction.
For ICD coding, this structure allows related codes to share output tokens while remaining distinguishable by unique identifiers. As a result, code prediction is broken down into a series of shared token decisions, allowing related codes to use portions of the same prediction space.

Medical codes contain rich semantic and structural information, including descriptions, official hierarchies, and relations between codes \citep{mullenbach-etal-2018-explainable,cao-etal-2020-hypercore,nguyen-etal-2023-two,luo-etal-2024-corelation}.
Recent work has used this information to construct medical representations. MedTok combines text and graph information in a shared token space \citep{su2025medtok}, SHARK uses residual $K$-means (RK-Means) to encode semantic factors among medical codes \citep{zhang2026shark}, and MedRQ incorporates medical ontologies into residual vector quantization \citep{lv2026medrq}. These studies motivate the use of semantic and structural information in medical code representations. Our focus is on Semantic IDs as direct generative targets for ICD coding. Models generate SIDs from clinical documents, and each complete SID maps uniquely to an ICD code. Beyond overall coding performance, we investigate two aspects of SID design: which ICD codes share SID tokens and how these tokens are organized into SID sequences. We examine how these choices affect downstream coding performance and the structural properties of the representations.

\section{Method}
\label{sec:method}

\begin{figure*}[t]
  \centering
  \includegraphics[width=\textwidth]{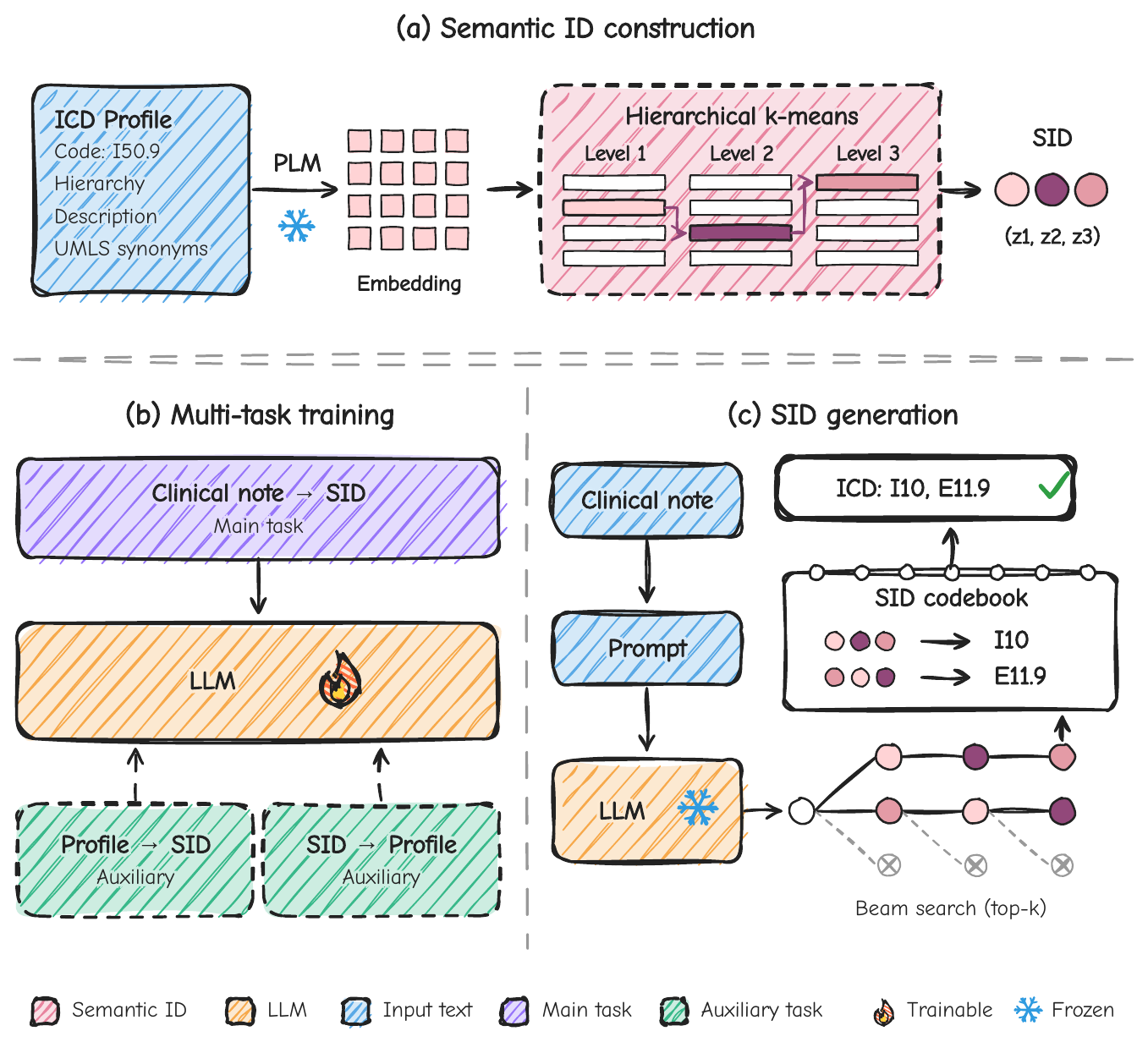}
  \caption{Overview of \icdsid{}. (a) ICD code profiles are encoded by the Qwen3 embedding model and clustered using hierarchical $K$-means to create Semantic IDs. (b) A language model learns to map clinical notes to SID sequences, along with auxiliary tasks for profile-to-SID and SID-to-profile predictions. (c) During inference, beam search generates SID sequences, which are then mapped back to ICD codes through the SID codebook.}
  \label{fig:overview}
\end{figure*}

As shown in Figure~\ref{fig:overview}, \icdsid{} performs generative ICD coding through three main stages. First, we create a SID codebook based on code profiles, where related codes can share prefixes, but each SID identifies a specific ICD code. Next, we train a language model to generate SID sequences from clinical documents, alongside bidirectional profile--SID auxiliary tasks. During inference, we decode to produce valid SID sequences, which are then mapped back to ICD codes using the codebook.

\subsection{Problem Formulation}
\label{sec:problem}

Let \( \codes \) represent the set of ICD codes. Given a clinical document \( x \), the goal of automatic ICD coding is to predict a set of codes \( Y \subseteq \codes \) that are supported by the content of the document.

Let $n=|Y|$ and denote a fixed canonical ordering of the target set
by $(c_{(1)},\ldots,c_{(n)})$. A generative ICD coder produces this sequence
autoregressively.
The generated codes are converted into a prediction set $\hat{Y}$ for evaluation.
Instead of generating ICD codes directly, \icdsid{} generates each code as a sequence of Semantic ID tokens.

\subsection{Semantic ID Construction}
\label{sec:SID}

For each ICD code $c$, we construct a code profile containing its hierarchy information, official description, and UMLS synonyms \citep{bodenreider2004umls} (see Appendix~\ref{sec:appendix-code-profiles} for details). We encode the profile with the Qwen3 embedding model \citep{zhang2025qwen3embedding} to obtain an embedding $\mathbf h_c$ and organize the embeddings using hierarchical $K$-means (HKM).
Starting from the code set, we recursively apply $K$-means within each parent group to form up to $K$ child groups.
The local cluster index of code $c$ at level $\ell$ is
\begin{equation}
  z_\ell(c)=\arg\min_j\|\mathbf h_c-\boldsymbol\mu_j\|_2^2,
  \label{eq:hkm-assignment}
\end{equation}
where $j$ indexes the child clusters within the current parent group and $\boldsymbol\mu_j$ is the corresponding centroid.
Thus, $z_\ell(c)$ is the index of the cluster whose centroid is closest to $\mathbf h_c$.
Repeating this process for $L$ levels yields the path $(z_1(c),\ldots,z_L(c))$.
Codes in the same group at level $\ell$ share the first $\ell$ path indices.

The Semantic ID of code \(c\), denoted by \(\phi(c)\), is the token sequence corresponding to this path.
We append a unique identifier (UID) token $u_c$ only when multiple codes share the same complete path, ensuring that each SID identifies an ICD code. The resulting one-to-one mapping between ICD codes and SIDs forms the SID codebook.
The SID codebook is constructed offline and held fixed throughout downstream training.

\begin{figure}[t]
  \centering
  \includegraphics[width=\columnwidth]{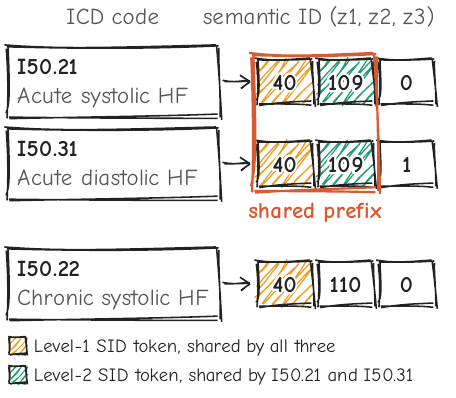}
  \caption{Shared SID prefixes and code-specific branches in the experimental codebook.
    HF denotes heart failure.}
  \label{fig:sid-worked-example}
\end{figure}

Figure~\ref{fig:sid-worked-example} shows three entries from the
codebook, illustrating how shared prefixes preserve
relatedness while distinguishing individual codes.
I50.21 and I50.31 describe acute systolic and acute diastolic heart
failure, respectively. They share the prefix $(40,109)$ and differ
at the third index, yielding paths $(40,109,0)$ and $(40,109,1)$.
I50.22, describing chronic systolic heart failure, shares the first
index but follows a different second-level branch, with path $(40,110,0)$.

\subsection{Multi-Task Training}
\label{sec:generative-adapter}

Each SID is represented as a short sequence of newly added vocabulary tokens, one for each hierarchy level, followed by an optional UID token.
Each SID token is determined by its hierarchy level and local cluster index.
For example, the paths for I50.21 and I50.31 illustrated in
Figure~\ref{fig:sid-worked-example} are represented as
\colorbox{gray!15}{$\langle \mathrm{SID}^{(1)}_{40}\rangle
 \langle \mathrm{SID}^{(2)}_{109}\rangle
 \langle \mathrm{SID}^{(3)}_{0}\rangle$}
and
\colorbox{gray!15}{$\langle \mathrm{SID}^{(1)}_{40}\rangle
 \langle \mathrm{SID}^{(2)}_{109}\rangle
 \langle \mathrm{SID}^{(3)}_{1}\rangle$}, respectively. They share the first two tokens and differ in the third.

Cluster indices are specific to each parent group and are reused across branches. Therefore, each SID token is interpreted in the context of its preceding prefix, with the semantic structure represented by a hierarchical path.

The SID codebook maintains the identity of each code while representing its predictions as a sequence of tokens. For a given code $c$, let $\phi(c) = (s_1, \ldots, s_m)$ denote its complete SID, where $s_i$
is the $i$-th SID token. The probability of this representation is factorized autoregressively as follows:
\begin{equation}
  p_\theta(\phi(c) \mid x) = \prod_{t=1}^{m} p_\theta(s_t \mid x, s_{<t}).
  \label{eq:sid-factorization}
\end{equation}
Codes that share a prefix also share early token targets, while subsequent tokens differentiate their identities.

For a document with multiple codes, we concatenate their SID representations in canonical code order
\begin{equation}
  s(Y)
  =
  \bigl[
    \phi(c_{(1)}),\ldots,\phi(c_{(n)})
  \bigr],
  \label{eq:sid-serialization}
\end{equation}
Here, $s(Y)$ denotes the serialized SID token sequence
for the code set $Y$, with adjacent SIDs separated by \colorbox{gray!15}{\texttt{<SEP>}}, and the sequence is terminated by \colorbox{gray!15}{\texttt{<END>}}.

The main task maps a clinical note to its SID sequence. We additionally use two auxiliary tasks, profile-to-SID and SID-to-profile, to align the learned SID tokens with code semantics. The profiles are those defined in Section~\ref{sec:SID}.
The SID codebook specifies the mapping of identifiers to codes, but this mapping is constructed offline and is not learned by the language model. Therefore, we implement bidirectional profile–SID prediction as a form of semantic alignment supervision.
For each code $c \in \codes$, we construct one profile-to-SID example and one SID-to-profile example using its fixed SID and code profile.
The prompt templates for all three tasks are provided in Appendix~\ref{sec:appendix-task-templates}.
We shuffle the examples within each task and construct each batch with examples from all three tasks.
Each training example is used once per epoch.
For input $x$ and target sequence $y$, the cross-entropy loss is
\begin{equation}
  \ell
  = -\sum_{t=1}^{|y|}
  \log p_{\theta}\bigl(y_t\mid x,y_{<t}\bigr).
  \label{eq:token-loss}
\end{equation}
For each task, we average the loss over all supervised target tokens.
The training objective is
\begingroup
\setlength{\medmuskip}{1mu}
\setlength{\thickmuskip}{2mu}
\begin{equation}
  \mathcal{L}
  = \lambda_1 \mathcal{L}_{\scriptscriptstyle\mathrm{note}\rightarrow\mathrm{SID}}
  + \lambda_2 \mathcal{L}_{\scriptscriptstyle\mathrm{profile}\rightarrow\mathrm{SID}}
  + \lambda_3 \mathcal{L}_{\scriptscriptstyle\mathrm{SID}\rightarrow\mathrm{profile}},
  \label{eq:generative-loss}
\end{equation}
\endgroup
where $(\lambda_1,\lambda_2,\lambda_3)=(0.8,0.1,0.1)$.
This normalization makes the task weights independent of the
number of target tokens in each task.

\subsection{SID Generation}
\label{sec:constrained-decoding}

During inference, the model generates a SID sequence from a clinical document using beam search.
Within each SID, only tokens that extend the current prefix toward a valid SID are allowed. If a UID token is required to identify the code, the SID is considered complete only after the UID token is generated. Once a complete SID is generated, the model may begin the next SID or terminate the sequence.

Each completed SID is mapped back to its unique ICD code through the SID codebook.
The recovered codes form the prediction set $\hat{Y}$.

\section{Experiments}
\label{sec:experiments}

We evaluate \icdsid{} through the following research questions:
\textbf{RQ1:} Does generating SIDs improve LLM-based ICD coding compared with directly generating original ICD codes?
\textbf{RQ2:} How do the components of \icdsid{} (i.e., the semantic assignment of SIDs and the two auxiliary tasks) affect the performance?
\textbf{RQ3:} How do different SID construction methods affect coding performance?

\subsection{Experimental Setup}
\label{sec:experimental-setup}

\subsubsection{Datasets}

We evaluate on the MIMIC-III ICD-9 dataset \citep{johnson2016mimiciii} and the MIMIC-IV ICD-10-CM dataset \citep{johnson2023mimiciv}, using the train, validation, and test splits from \citet{edin-etal-2023-automated}. We retain all labels in each processed dataset. Table~\ref{tab:dataset-statistics} summarizes the dataset statistics.

\begin{table}[htbp]
  \centering
  \small
  \setlength{\tabcolsep}{4pt}
  \begin{tabular*}{\columnwidth}{@{\extracolsep{\fill}}lcc@{}}
    \toprule
    & \textbf{MIMIC-III ICD-9} & \textbf{MIMIC-IV ICD-10-CM} \\
    \midrule
    \#Docs & 52,723 & 122,304 \\
    \#Patients & 41,126 & 65,675 \\
    \#Codes & 8,929 & 16,155 \\
    \bottomrule
  \end{tabular*}
  \caption{Dataset statistics for MIMIC-III v1.4 and MIMIC-IV v2.2.}
  \label{tab:dataset-statistics}
\end{table}

\begin{table*}[t]
  \centering
  \small
  \setlength{\tabcolsep}{2.5pt}
  \renewcommand{\arraystretch}{1.12}
  \begin{tabular*}{\textwidth}{@{\extracolsep{\fill}}cccccccccc@{}}
    \toprule
    &
      & \multicolumn{4}{c}{MIMIC-III ICD-9}
      & \multicolumn{4}{c}{MIMIC-IV ICD-10-CM} \\
    \cmidrule(lr){3-6} \cmidrule(lr){7-10}
    Backbone & Method & Micro-F1 & Macro-F1 & Precision & Recall
      & Micro-F1 & Macro-F1 & Precision & Recall \\
    \midrule
    Qwen3-0.6B & Raw
      & 50.15\textsubscript{$\pm$0.47} & 5.64\textsubscript{$\pm$0.16} & 50.65\textsubscript{$\pm$0.48} & 49.65\textsubscript{$\pm$0.50} & 53.40\textsubscript{$\pm$0.42} & 6.29\textsubscript{$\pm$0.24} & 53.38\textsubscript{$\pm$0.38} & 53.43\textsubscript{$\pm$0.44} \\
      & \icdsid{}
      & \textbf{53.65}\textsubscript{$\pm$0.21} & \textbf{7.14}\textsubscript{$\pm$0.27} & 55.29\textsubscript{$\pm$0.51} & 52.10\textsubscript{$\pm$0.59} & \textbf{56.35}\textsubscript{$\pm$0.20} & \textbf{8.10}\textsubscript{$\pm$0.15} & 58.53\textsubscript{$\pm$0.36} & 54.33\textsubscript{$\pm$0.35} \\
    \midrule
    Qwen3-4B & Raw
      & 55.41\textsubscript{$\pm$0.57} & 10.12\textsubscript{$\pm$0.28} & 55.13\textsubscript{$\pm$0.48} & 55.69\textsubscript{$\pm$0.40} & 58.17\textsubscript{$\pm$0.25} & 10.96\textsubscript{$\pm$0.28} & 59.56\textsubscript{$\pm$0.42} & 56.85\textsubscript{$\pm$0.34} \\
      & \icdsid{}
      & \textbf{57.69}\textsubscript{$\pm$0.21} & \textbf{11.50}\textsubscript{$\pm$0.21} & 59.22\textsubscript{$\pm$0.47} & 56.23\textsubscript{$\pm$0.36} & \textbf{59.51}\textsubscript{$\pm$0.43} & \textbf{12.17}\textsubscript{$\pm$0.21} & 61.35\textsubscript{$\pm$0.36} & 57.77\textsubscript{$\pm$0.37} \\
    \midrule
    Gemma3-1B & Raw
      & 49.16\textsubscript{$\pm$0.43} & 6.26\textsubscript{$\pm$0.12} & 48.45\textsubscript{$\pm$0.53} & 49.89\textsubscript{$\pm$0.57} & 48.54\textsubscript{$\pm$0.40} & 4.14\textsubscript{$\pm$0.43} & 46.60\textsubscript{$\pm$0.34} & 50.65\textsubscript{$\pm$0.42} \\
      & \icdsid{}
      & \textbf{52.96}\textsubscript{$\pm$0.22} & \textbf{7.63}\textsubscript{$\pm$0.28} & 54.68\textsubscript{$\pm$0.31} & 51.35\textsubscript{$\pm$0.21} & \textbf{56.54}\textsubscript{$\pm$0.25} & \textbf{7.29}\textsubscript{$\pm$0.21} & 57.63\textsubscript{$\pm$0.20} & 55.50\textsubscript{$\pm$0.44} \\
    \midrule
    Gemma3-4B & Raw
      & 54.08\textsubscript{$\pm$0.39} & 9.79\textsubscript{$\pm$0.24} & 53.10\textsubscript{$\pm$0.32} & 55.09\textsubscript{$\pm$0.23} & 57.14\textsubscript{$\pm$0.26} & 10.74\textsubscript{$\pm$0.22} & 57.49\textsubscript{$\pm$0.40} & 56.79\textsubscript{$\pm$0.21} \\
      & \icdsid{}
      & \textbf{56.58}\textsubscript{$\pm$0.45} & \textbf{11.67}\textsubscript{$\pm$0.15} & 57.06\textsubscript{$\pm$0.57} & 56.10\textsubscript{$\pm$0.22} & \textbf{58.87}\textsubscript{$\pm$0.41} & \textbf{11.98}\textsubscript{$\pm$0.18} & 59.97\textsubscript{$\pm$0.28} & 57.81\textsubscript{$\pm$0.24} \\
    \bottomrule
  \end{tabular*}
  \caption{Generative ICD coding results on MIMIC-III ICD-9 and MIMIC-IV ICD-10-CM.
    Raw generates ICD codes directly, whereas \icdsid{} generates SIDs and maps them back to ICD codes.
    Scores are percentages; precision and recall are micro-averaged.
    Subscripts indicate 95\% confidence interval half-widths.}
  \label{tab:main_results}
\end{table*}

\subsubsection{Evaluation Metrics}
\label{sec:evaluation-metrics}

Following prior work on automatic ICD coding
\citep{mullenbach-etal-2018-explainable, vu2020label, huang-etal-2022-plm, yang2023multilabel},
we report Micro-F1 and Macro-F1 as the main evaluation metrics.
We use Micro-F1 as the primary metric and Macro-F1 to account for performance across individual codes.
We also report micro-averaged precision and recall.
All metrics are computed on the final predicted ICD code sets using exact code matching.

We report means across five random seeds, with 95\% confidence interval half-widths for each metric in Table~\ref{tab:main_results}.

\subsubsection{Baselines}

We utilize Qwen3 \citep{yang2025qwen3} and Gemma3 \citep{gemmateam2025gemma3} as our backbone families. In our main experiments, we focus on Qwen3-0.6B and Gemma3-1B, while also including Qwen3-4B and Gemma3-4B to evaluate larger-scale models. For each backbone model, \textbf{Raw} directly generates ICD codes in their original form, whereas \icdsid{} produces Semantic IDs that are subsequently mapped back to ICD codes.

\subsubsection{Implementation Details}
\label{sec:implementation-details}

Semantic IDs are built from Qwen3 embeddings using hierarchical $K$-means with a branching factor of $K=128$ and a depth of $L=3$, selected on the validation set (Appendix~\ref{sec:appendix-hyperparameters}).

We fine-tune all models for up to 6 epochs using AdamW with a learning rate of $3\times10^{-5}$ and set a maximum input length of 8,192 tokens. Fine-tuning was performed with Unsloth \citep{han2023unsloth} on a single NVIDIA RTX PRO 6000 GPU. We evaluate validation Micro-F1 after each epoch and stop training if it does not improve for two consecutive epochs. We select the checkpoint with the highest validation Micro-F1. Raw and \icdsid{} share the same data splits, backbone models, optimization settings, and checkpoint selection criteria. During inference, both use beam search with beam size 4. Raw generates valid ICD codes, while \icdsid{} generates valid Semantic ID sequences, which are then mapped back to ICD codes.

\subsection{Overall Performance (RQ1)}
\label{sec:results}
Table~\ref{tab:main_results} demonstrates that \icdsid{} outperforms direct ICD code generation. The Micro-F1 scores show an increase from 2.28 to 3.80 points on the MIMIC-III dataset and from 1.34 to 8.00 points on the MIMIC-IV dataset. These enhancements are observed for both Qwen3 and Gemma3 models across the two different sizes within each model family.

The gains in Micro-F1 scores are more significant for the smaller backbone models within each family across both datasets. For instance, on the MIMIC-IV dataset, \icdsid{} improves the performance of Gemma3-1B from 48.54 to 56.54, an improvement of 8.00 points. Gemma3-4B improves by 1.73 points. This trend is consistent with Qwen3 and also applies to the MIMIC-III dataset. The 4B models also show improvements, with increases ranging from 1.34 to 2.50 points across the two datasets.

The Macro-F1 score also increases across both datasets, rising by 1.37 to 1.88 points on MIMIC-III and by 1.21 to 3.15 points on MIMIC-IV. This improvement affects the metric that treats each code equally. Both precision and recall increase in every comparison, with precision showing larger absolute gains than recall. Consequently, the observed gains in the Micro-F1 score are accompanied by improved precision without any reduction in recall.

\subsection{Performance across Code Frequencies}
\label{sec:code-frequency}

We next examine how the performance of \icdsid{} varies with ICD code frequency. We categorize the ICD codes by their frequency in the training dataset and report the Micro-F1 scores for each category. Table~\ref{tab:code-frequency} indicates that \icdsid{} outperforms Raw across all five frequency groups.

The gains are most significant for codes that appear between 11 and 500 times, with increases ranging from 3.02 to 4.07 points. In contrast, codes that are observed more than 500 times show a gain of only 2.49 points. The greatest improvement is observed in the 101–500 group. For codes that are observed only 1 to 10 times, the Micro-F1 score rises from 3.53 to 4.15. Although performance remains low for both methods in rare codes, improvements are still evident.

Overall, the strongest gains are seen in the medium-frequency range, while the extreme tail of rare codes continues to present challenges.

\begin{table}[!htb]
  \centering
  \small
  \setlength{\tabcolsep}{7pt}
  \begin{tabular}{@{}lrccc@{}}
    \toprule
    Train freq. & \#Codes & Raw & \icdsid & $\Delta$ \\
    \midrule
    $>500$ & 444 & 60.75 & 63.24 & +2.49 \\
    101--500 & 887 & 37.66 & 41.73 & \textbf{+4.07} \\
    51--100 & 760 & 25.33 & 29.12 & +3.79 \\
    11--50 & 2,711 & 16.01 & 19.03 & +3.02 \\
    1--10 & 11,279 & 3.53 & 4.15 & +0.62 \\
    \bottomrule
  \end{tabular}
  \caption{Micro-F1 by training-set code frequency on MIMIC-IV
    ICD-10-CM using Qwen3-0.6B. \#Codes is the number of distinct codes
    in each group. $\Delta$ denotes the improvement. Bold marks the largest gain.}
  \label{tab:code-frequency}
\end{table}

\subsection{Ablation Study (RQ2)}
\label{sec:representation-ablations}

We first examine whether the gains of \icdsid{} depend on semantic SID assignment and token sharing. Random SID randomly reassigns the original SID sequences across ICD codes, preserving the SID structure while disrupting semantic relationships. Atomic ID assigns a unique token to each ICD code, eliminating token sharing across codes.

\begin{table}[!htb]
  \centering
  \small
  \setlength{\tabcolsep}{8pt}
  \begin{tabular}{@{}lll@{}}
    \toprule
    Variant & Micro-F1 & Macro-F1 \\
    \midrule
    \textbf{\icdsid{} (full)} & \textbf{56.35} & \textbf{8.10} \\
    \midrule
    \multicolumn{3}{@{}l}{\textit{Output representation}} \\
    \quad Atomic ID & 54.00\dn{2.35} & 6.92\dn{1.18} \\
    \quad Random SID & 53.74\dn{2.61} & 6.44\dn{1.66} \\
    \midrule
    \multicolumn{3}{@{}l}{\textit{Auxiliary tasks}} \\
    \quad w/o Profile$\rightarrow$SID & 55.22\dn{1.13} & 7.45\dn{0.65} \\
    \quad w/o SID$\rightarrow$Profile & 55.91\dn{0.44} & 7.86\dn{0.24} \\
    \quad w/o both & 54.30\dn{2.05} & 7.14\dn{0.96} \\
    \bottomrule
  \end{tabular}
  \caption{Ablations of the output representation and the auxiliary tasks on MIMIC-IV ICD-10-CM using Qwen3-0.6B.}
  \label{tab:representation-ablations}
\end{table}

As shown in Table~\ref{tab:representation-ablations}, Random SID and Atomic ID fall below \icdsid{} by 2.61 and 2.35 Micro-F1 points, respectively. The lower performance of Random SID indicates that simply having shared output tokens is insufficient. The way these tokens are assigned to ICD codes also plays a crucial role in performance.

We next evaluate the impact of the two auxiliary objectives while keeping the semantic SID codebook and the main task unchanged. Removing both objectives reduces Micro-F1 from 56.35 to 54.30 and Macro-F1 from 8.10 to 7.14. The removal of the profile-to-SID objective results in a greater decline than the removal of the SID-to-profile objective, with drops of 1.13 and 0.44 Micro-F1 points, and 0.65 and 0.24 Macro-F1 points, respectively. This indicates that both auxiliary objectives contribute to performance, with profile-to-SID having the more substantial individual effect.

\subsection{Effect of SID Construction (RQ3)}
\label{sec:sid-structure}

We compare HKM, RK-Means, and
the residual-quantized variational autoencoder (RQ-VAE) \citep{lee2022rqvae} in terms of
performance and the structural properties of their SIDs.
The constructions use the same ICD-10-CM and
embeddings, with $K=128$ and $L=3$.
Table~\ref{tab:sid-construction-performance} reports
results on MIMIC-IV ICD-10-CM using Qwen3-0.6B.
HKM achieves the highest Micro-F1 of 56.35, exceeding
RK-Means by 0.93 points and RQ-VAE by 2.39 points.
Macro-F1 follows the same rank, with scores of 8.10,
7.54, and 6.41 for HKM, RK-Means, and RQ-VAE,
respectively.

\begin{table}[!htb]
  \centering
  \small
  \setlength{\tabcolsep}{10pt}
  \begin{tabular}{@{}lcc@{}}
    \toprule
    SID construction & Micro-F1 & Macro-F1 \\
    \midrule
    HKM & \textbf{56.35} & \textbf{8.10} \\
    RK-Means & 55.42 & 7.54 \\
    RQ-VAE & 53.96 & 6.41 \\
    \bottomrule
  \end{tabular}
  \caption{ICD coding performance of different SID constructions on
    MIMIC-IV ICD-10-CM using Qwen3-0.6B. Bold indicates the highest F1.}
  \label{tab:sid-construction-performance}
\end{table}

\begin{figure}[t]
  \centering
  \includegraphics[width=\columnwidth]{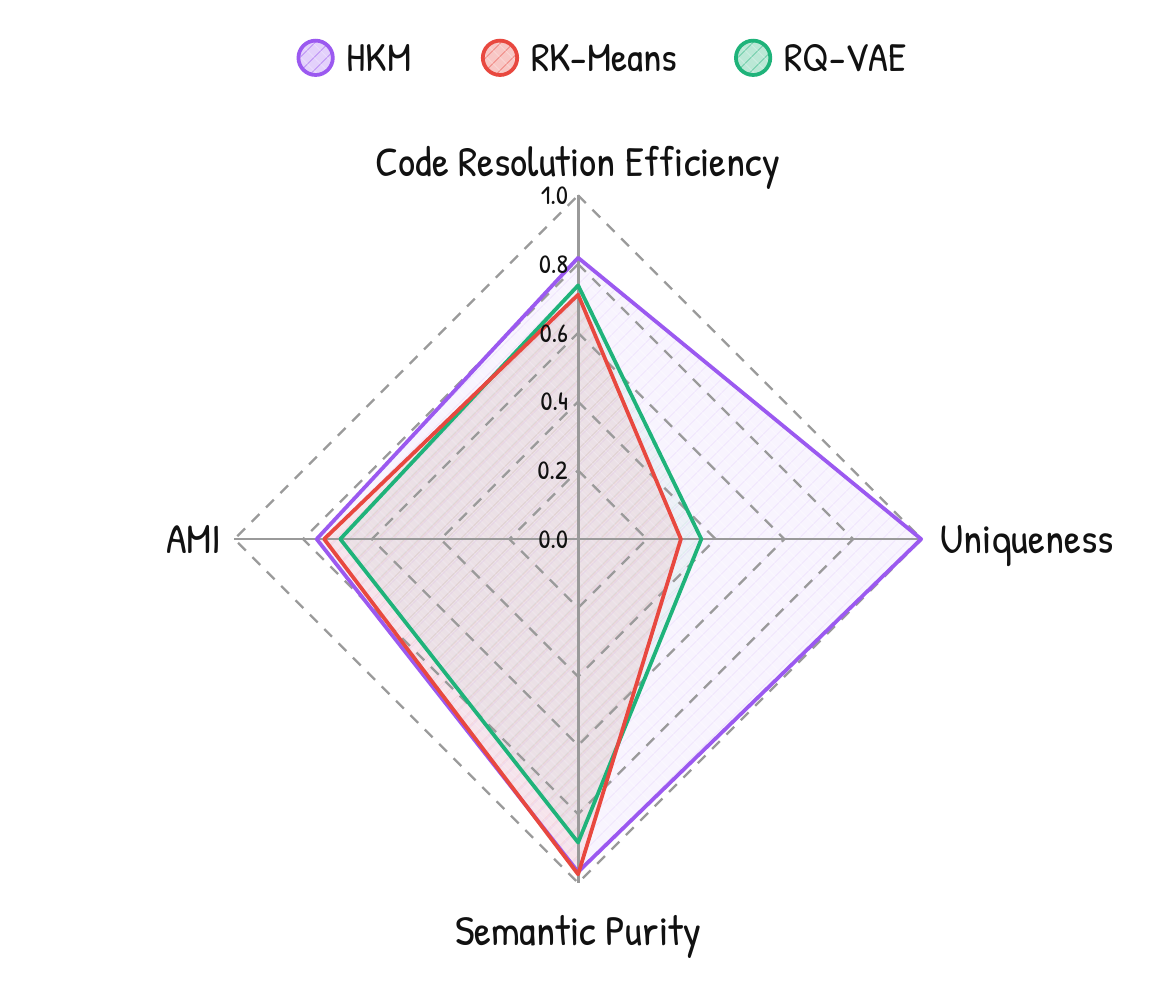}
  \caption{Structural properties of native SID paths in the ICD-10-CM. Semantic Purity refers to chapter taxonomy purity; RQ-VAE is assessed using native nearest-neighbor assignments.}
  \label{fig:sid-construction-radar}
\end{figure}

Figure~\ref{fig:sid-construction-radar} provides a comparison of the structures created by the three construction methods. We evaluate the native SID paths before disambiguation along two dimensions. The first dimension, taxonomy alignment, is assessed using Adjusted Mutual Information (AMI) and  Semantic Purity, which measure how well SID groups correspond to the ICD taxonomy. The second dimension measures how well SID paths distinguish individual ICD codes, using full-path Uniqueness and Code Resolution Efficiency (CRE). Uniqueness measures the fraction of codes with a unique complete SID path, while CRE measures how progressively codes are distinguished as the SID path becomes longer. Formal definitions can be found in Appendix~\ref{sec:appendix-sid-metrics}.

HKM and RK-Means demonstrate similar levels of taxonomy alignment, while RQ-VAE scores lower on both alignment measures. The methods differ significantly in code separation. HKM generates unique native paths for 99.79\% of codes, compared to 29.90\% for RK-Means and 35.85\% for RQ-VAE. Additionally, HKM achieves the highest CRE at 0.818, followed by RQ-VAE at 0.738 and RK-Means at 0.711. 

However, stronger path separation does not fully account for the performance differences: RQ-VAE has higher Uniqueness and CRE than RK-Means, yet it achieves lower performance. Therefore, no single structural metric fully accounts for the differences in performance.

\section{Discussion}
\label{sec:discussion}

A key design decision in generative ICD coding is how prediction decisions are shared across different codes. The results from the assignment ablation in Table~\ref{tab:representation-ablations} indicate that this sharing is influenced by the relationship between SIDs and ICD codes. Random SID yields lower coding performance. This suggests that the gains arise not only from the structure of SIDs but also from assigning semantically related codes to shared prefixes.

The auxiliary-task ablation provides additional evidence: when supervision is removed from either profile-to-SID or SID-to-profile, performance decreases, with a more pronounced drop when supervision on profile-to-SID is omitted. Collectively, these results demonstrate that both the semantic assignment of SIDs and the explicit alignment between profiles and SIDs play vital roles in enhancing the representation's effectiveness.

The frequency analysis shows that SemICD improves performance across all evaluated frequency groups, with the largest gains for codes observed 11--500 times during training (Table~\ref{tab:code-frequency}). The improvement is much smaller for codes observed only 1--10 times, indicating that extremely sparse labels remain difficult for both methods. These results suggest that shared output structure is most useful when supervision is limited but not extremely sparse. The construction comparison further suggests that how this sharing is organized also matters. Related codes can share early prediction decisions, while later decisions must distinguish individual codes. HKM implements this pattern through hierarchical refinement, with shared structure at coarse levels and progressively finer distinctions at later levels.

HKM demonstrates the strongest performance while maintaining taxonomy alignment, similar to RK-Means, and achieving higher uniqueness and CRE. However, these properties do not solely account for the differences in performance. RQ-VAE achieves greater path uniqueness and CRE than RK-Means, but it has a lower  F1 score.

These comparisons indicate that taxonomy alignment and path separation represent complementary aspects of SID representations, but neither alone determines performance. Therefore, it is essential to evaluate structural metrics alongside assessments when analyzing SID constructions.

\section{Conclusion}
\label{sec:conclusion}

We presented \icdsid{}, a framework designed for generative ICD coding using Semantic ID. The \icdsid{} framework combines SID prediction, auxiliary code-profile tasks, and prefix-constrained decoding. When evaluated on the MIMIC-III and MIMIC-IV datasets, \icdsid{} demonstrates improved performance over direct ICD code generation. Our ablation study reveals that the performance is influenced by how SID sequences are assigned to ICD codes. Furthermore, a comparison of construction strategies indicates that the HKM method performs best among the evaluated SID construction techniques. These findings emphasize that both the assignment of Semantic IDs and their construction are crucial to the effectiveness of generative ICD coding.

\section*{Limitations}

Our study is limited to MIMIC-III and MIMIC-IV within the evaluated diagnosis coding settings. As a result, the findings may not be directly applicable to other institutions, note types, or editions of the ICD. Additionally, the SID codebook is created offline from code-profile embeddings and remains fixed throughout downstream training. This means the induced sharing structure is not optimized specifically for the document-to-code objective. Finally, the generation of SID is autoregressive and prefix-conditioned. If a mistake is made at an early level of the SID, it can limit subsequent predictions to the incorrect branch. Although constrained decoding ensures that the generated path is valid, it does not guarantee that an early branching error can be corrected.

\bibliography{custom}

@inproceedings{mullenbach-etal-2018-explainable,
  title = {Explainable Prediction of Medical Codes from Clinical Text},
  author = {Mullenbach, James and Wiegreffe, Sarah and Duke, Jon and Sun, Jimeng and Eisenstein, Jacob},
  booktitle = {Proceedings of the 2018 Conference of the North American Chapter of the Association for Computational Linguistics: Human Language Technologies, Volume 1 (Long Papers)},
  year = {2018},
  pages = {1101--1111},
  publisher = {Association for Computational Linguistics},
  doi = {10.18653/v1/N18-1100},
  url = {https://aclanthology.org/N18-1100/}
}

@inproceedings{edin-etal-2023-automated,
  title = {Automated Medical Coding on {MIMIC-III} and {MIMIC-IV}: A Critical Review and Replicability Study},
  author = {Edin, Joakim and Junge, Alexander and Havtorn, Jakob D. and Borgholt, Lasse and Maistro, Maria and Ruotsalo, Tuukka and Maal{\o}e, Lars},
  booktitle = {Proceedings of the 46th International ACM SIGIR Conference on Research and Development in Information Retrieval},
  year = {2023},
  pages = {2572--2582},
  publisher = {Association for Computing Machinery},
  doi = {10.1145/3539618.3591918},
  url = {https://doi.org/10.1145/3539618.3591918}
}

@inproceedings{vu2020label,
  title = {A Label Attention Model for {ICD} Coding from Clinical Text},
  author = {Vu, Thanh and Nguyen, Dat Quoc and Nguyen, Anthony N.},
  booktitle = {Proceedings of the Twenty-Ninth International Joint Conference on Artificial Intelligence},
  year = {2020},
  pages = {3335--3341},
  publisher = {International Joint Conferences on Artificial Intelligence Organization},
  doi = {10.24963/IJCAI.2020/461},
  url = {https://doi.org/10.24963/ijcai.2020/461}
}

@inproceedings{huang-etal-2022-plm,
  title = {{PLM}-{ICD}: Automatic {ICD} Coding with Pretrained Language Models},
  author = {Huang, Chao-Wei and Tsai, Shang-Chi and Chen, Yun-Nung},
  booktitle = {Proceedings of the 4th Clinical Natural Language Processing Workshop},
  year = {2022},
  pages = {10--20},
  publisher = {Association for Computational Linguistics},
  doi = {10.18653/v1/2022.clinicalnlp-1.2},
  url = {https://aclanthology.org/2022.clinicalnlp-1.2/}
}

@article{yang2023multilabel,
  title = {Multi-Label Few-Shot {ICD} Coding as Autoregressive Generation with Prompt},
  author = {Yang, Zhichao and Kwon, Sunjae and Yao, Zonghai and Yu, Hong},
  journal = {Proceedings of the AAAI Conference on Artificial Intelligence},
  year = {2023},
  volume = {37},
  number = {4},
  pages = {5366--5374},
  doi = {10.1609/aaai.v37i4.25668},
  url = {https://doi.org/10.1609/aaai.v37i4.25668}
}

@inproceedings{baksi-etal-2025-medcoder,
  title = {{M}ed{C}od{ER}: A Generative {AI} Assistant for Medical Coding},
  author = {Baksi, Krishanu Das and Soba, Elijah and Higgins, John J and Saini, Ravi and Wood, Jaden and Cook, Jane and Scott, Jack I and Pudota, Nirmala and Weninger, Tim and Bowen, Edward and Bhattacharya, Sanmitra},
  booktitle = {Proceedings of the 2025 Conference of the Nations of the Americas Chapter of the Association for Computational Linguistics: Human Language Technologies (Volume 3: Industry Track)},
  year = {2025},
  pages = {449--459},
  publisher = {Association for Computational Linguistics},
  doi = {10.18653/v1/2025.naacl-industry.37},
  url = {https://aclanthology.org/2025.naacl-industry.37/}
}

@inproceedings{cao-etal-2020-hypercore,
  title = {{H}yper{C}ore: Hyperbolic and Co-graph Representation for Automatic {ICD} Coding},
  author = {Cao, Pengfei and Chen, Yubo and Liu, Kang and Zhao, Jun and Liu, Shengping and Chong, Weifeng},
  booktitle = {Proceedings of the 58th Annual Meeting of the Association for Computational Linguistics},
  year = {2020},
  pages = {3105--3114},
  publisher = {Association for Computational Linguistics},
  doi = {10.18653/v1/2020.acl-main.282},
  url = {https://aclanthology.org/2020.acl-main.282/}
}

@inproceedings{nguyen-etal-2023-two,
  title = {A Two-Stage Decoder for Efficient {ICD} Coding},
  author = {Nguyen, Thanh-Tung and Schlegel, Viktor and Ramesh Kashyap, Abhinav and Winkler, Stefan},
  booktitle = {Findings of the Association for Computational Linguistics: ACL 2023},
  year = {2023},
  pages = {4658--4665},
  publisher = {Association for Computational Linguistics},
  doi = {10.18653/v1/2023.findings-acl.285},
  url = {https://aclanthology.org/2023.findings-acl.285/}
}

@inproceedings{luo-etal-2024-corelation,
  title = {{C}o{R}elation: Boosting Automatic {ICD} Coding through Contextualized Code Relation Learning},
  author = {Luo, Junyu and Wang, Xiaochen and Wang, Jiaqi and Chang, Aofei and Wang, Yaqing and Ma, Fenglong},
  booktitle = {Proceedings of the 2024 Joint International Conference on Computational Linguistics, Language Resources and Evaluation (LREC-COLING 2024)},
  year = {2024},
  pages = {3997--4007},
  publisher = {ELRA and ICCL},
  url = {https://aclanthology.org/2024.lrec-main.355/}
}

@inproceedings{tay2022transformer,
  title = {Transformer Memory as a Differentiable Search Index},
  author = {Tay, Yi and Tran, Vinh and Dehghani, Mostafa and Ni, Jianmo and Bahri, Dara and Mehta, Harsh and Qin, Zhen and Hui, Kai and Zhao, Zhe and Gupta, Jai Prakash and Schuster, Tal and Cohen, William W. and Metzler, Donald},
  booktitle = {Advances in Neural Information Processing Systems 35},
  year = {2022},
  pages = {21831--21843},
  url = {https://proceedings.neurips.cc/paper_files/paper/2022/hash/892840a6123b5ec99ebaab8be1530fba-Abstract-Conference.html}
}

@inproceedings{zheng2024adapting,
  title = {Adapting Large Language Models by Integrating Collaborative Semantics for Recommendation},
  author = {Zheng, Bowen and Hou, Yupeng and Lu, Hongyu and Chen, Yu and Zhao, Wayne Xin and Chen, Ming and Wen, Ji-Rong},
  booktitle = {2024 IEEE 40th International Conference on Data Engineering (ICDE)},
  year = {2024},
  pages = {1435--1448},
  publisher = {IEEE},
  doi = {10.1109/ICDE60146.2024.00118},
  url = {https://doi.org/10.1109/ICDE60146.2024.00118}
}

@misc{luo2024qarm,
  title = {{QARM}: Quantitative Alignment Multi-Modal Recommendation at Kuaishou},
  author = {Luo, Xinchen and Cao, Jiangxia and Sun, Tianyu and Yu, Jinkai and Huang, Rui and Yuan, Wei and Lin, Hezheng and Zheng, Yichen and Wang, Shiyao and Hu, Qigen and Qiu, Changqing and Zhang, Jiaqi and Zhang, Xu and Yan, Zhiheng and Zhang, Jingming and Zhang, Simin and Wen, Mingxing and Liu, Zhaojie and Gai, Kun and Zhou, Guorui},
  year = {2024},
  eprint = {2411.11739},
  archivePrefix = {arXiv},
  primaryClass = {cs.IR},
  url = {https://arxiv.org/abs/2411.11739}
}

@inproceedings{rajput2023recommender,
  title = {Recommender Systems with Generative Retrieval},
  author = {Rajput, Shashank and Mehta, Nikhil and Singh, Anima and Keshavan, Raghunandan Hulikal and Vu, Trung and Heldt, Lukasz and Hong, Lichan and Tay, Yi and Tran, Vinh Q. and Samost, Jonah and Kula, Maciej and Chi, Ed H. and Sathiamoorthy, Maheswaran},
  booktitle = {Advances in Neural Information Processing Systems 36},
  year = {2023},
  pages = {10299--10315},
  doi = {10.52202/075280-0452},
  url = {https://papers.neurips.cc/paper_files/paper/2023/hash/20dcab0f14046a5c6b02b61da9f13229-Abstract-Conference.html}
}

@inproceedings{su2025medtok,
  title = {Multimodal Medical Code Tokenizer},
  author = {Su, Xiaorui and Messica, Shvat and Huang, Yepeng and Johnson, Ruth and Fesser, Lukas and Gao, Shanghua and Sahneh, Faryad and Zitnik, Marinka},
  booktitle = {Proceedings of the 42nd International Conference on Machine Learning},
  year = {2025},
  volume = {267},
  series = {Proceedings of Machine Learning Research},
  pages = {57030--57045},
  publisher = {PMLR},
  url = {https://proceedings.mlr.press/v267/su25b.html}
}

@inproceedings{zhang2026shark,
  title = {{SHARK}: Modeling Semantic Hierarchy of Medical Code via Residual {K}-Means Quantization},
  author = {Zhang, Kaisong and Lv, Hang and Tan, Yanchao and Lin, Zhigang and Zhang, Hengyu and Chen, Xiping and Yang, Carl},
  booktitle = {ICASSP 2026--2026 IEEE International Conference on Acoustics, Speech and Signal Processing (ICASSP)},
  year = {2026},
  pages = {12832--12836},
  publisher = {IEEE},
  doi = {10.1109/ICASSP55912.2026.11462768},
  url = {https://doi.org/10.1109/ICASSP55912.2026.11462768}
}

@inproceedings{lv2026medrq,
  title = {Towards Efficient and Interpretable Medical Concept Representation via Ontology-Driven Residual Vector Quantization},
  author = {Lv, Hang and Zhang, Kaisong and Tan, Yanchao and Chen, Xing},
  booktitle = {Proceedings of the ACM Web Conference 2026},
  year = {2026},
  pages = {8645--8648},
  publisher = {ACM},
  doi = {10.1145/3774904.3792930},
  url = {https://doi.org/10.1145/3774904.3792930}
}

@article{wang2026posttraining,
  title = {Can Post-Training Turn {LLMs} into Good Medical Coders? An Empirical Study of Generative {ICD} Coding},
  author = {Wang, Ziqing and Li, Weihao and Chen, Shijie and Luo, Yuan and Ding, Kaize},
  journal = {arXiv preprint arXiv:2606.13940},
  year = {2026},
  doi = {10.48550/arXiv.2606.13940},
  url = {https://arxiv.org/abs/2606.13940}
}

@article{yang2025qwen3,
  title = {{Qwen3} Technical Report},
  author = {Yang, An and others},
  journal = {arXiv preprint arXiv:2505.09388},
  year = {2025},
  doi = {10.48550/arXiv.2505.09388},
  url = {https://arxiv.org/abs/2505.09388}
}

@article{gemmateam2025gemma3,
  title = {{Gemma 3} Technical Report},
  author = {{Gemma Team} and others},
  journal = {arXiv preprint arXiv:2503.19786},
  year = {2025},
  doi = {10.48550/arXiv.2503.19786},
  url = {https://arxiv.org/abs/2503.19786}
}

@article{zhang2025qwen3embedding,
  title = {{Qwen3 Embedding}: Advancing Text Embedding and Reranking Through Foundation Models},
  author = {Zhang, Yanzhao and others},
  journal = {arXiv preprint arXiv:2506.05176},
  year = {2025},
  doi = {10.48550/arXiv.2506.05176},
  url = {https://arxiv.org/abs/2506.05176}
}

@misc{han2023unsloth,
  title = {Unsloth},
  author = {Han, Daniel and Han, Michael and {Unsloth Team}},
  year = {2023},
  url = {https://github.com/unslothai/unsloth}
}

@article{johnson2016mimiciii,
  title = {{MIMIC-III Clinical Database}},
  author = {Johnson, Alistair and Pollard, Tom and Mark, Roger},
  journal = {PhysioNet},
  year = {2016},
  note = {Version 1.4},
  doi = {10.13026/C2XW26},
  url = {https://doi.org/10.13026/C2XW26}
}

@article{johnson2023mimiciv,
  title = {{MIMIC-IV}},
  author = {Johnson, Alistair and Bulgarelli, Lucas and Pollard, Tom and Horng, Steven and Celi, Leo Anthony and Mark, Roger},
  journal = {PhysioNet},
  year = {2023},
  note = {Version 2.2},
  doi = {10.13026/6mm1-ek67},
  url = {https://doi.org/10.13026/6mm1-ek67}
}

@article{bodenreider2004umls,
  title = {The {Unified Medical Language System} ({UMLS}): Integrating Biomedical Terminology},
  author = {Bodenreider, Olivier},
  journal = {Nucleic Acids Research},
  volume = {32},
  number = {Database issue},
  pages = {D267--D270},
  year = {2004},
  doi = {10.1093/nar/gkh061},
  url = {https://doi.org/10.1093/nar/gkh061}
}

@inproceedings{lee2022rqvae,
  title = {Autoregressive Image Generation Using Residual Quantization},
  author = {Lee, Doyup and Kim, Chiheon and Kim, Saehoon and Cho, Minsu and Han, Wook-Shin},
  booktitle = {Proceedings of the IEEE/CVF Conference on Computer Vision and Pattern Recognition},
  year = {2022},
  pages = {11523--11532},
  url = {https://openaccess.thecvf.com/content/CVPR2022/html/Lee_Autoregressive_Image_Generation_Using_Residual_Quantization_CVPR_2022_paper.html}
}

@article{falis2024gpt,
  title = {Can {GPT}-3.5 Generate and Code Discharge Summaries?},
  author = {Falis, Mat{\'u}{\v{s}} and Gema, Aryo Pradipta and Dong, Hang and Daines, Luke and Basetti, Siddharth and Holder, Michael and Penfold, Rose S. and Birch, Alexandra and Alex, Beatrice},
  journal = {Journal of the American Medical Informatics Association},
  year = {2024},
  volume = {31},
  number = {10},
  pages = {2284--2293},
  doi = {10.1093/jamia/ocae132},
  url = {https://doi.org/10.1093/jamia/ocae132}
}

@inproceedings{yuan-etal-2025-toward,
  title = {Toward Reliable Clinical Coding with Language Models: Verification and Lightweight Adaptation},
  author = {Yuan, Moy and Shing, Han-Chin and Strong, Mitch and Shivade, Chaitanya},
  booktitle = {Proceedings of the 2025 Conference on Empirical Methods in Natural Language Processing: Industry Track},
  year = {2025},
  pages = {173--184},
  publisher = {Association for Computational Linguistics},
  doi = {10.18653/v1/2025.emnlp-industry.12},
  url = {https://aclanthology.org/2025.emnlp-industry.12/}
}

@article{chen2026krel,
  title = {{KREL}: Automatic Medical Coding via Knowledge-Guided Reasoning over Clinical Evidence with {LLMs}},
  author = {Chen, Xubin and Zhou, Yipeng and Sun, Wen and Huang, Chengkai and Fu, Xiaoming and Sheng, Quan Z.},
  journal = {arXiv preprint arXiv:2608.20887},
  year = {2026},
  doi = {10.48550/arXiv.2608.20887},
  url = {https://arxiv.org/abs/2608.20887}
}

@inproceedings{motzfeldt-etal-2025-code,
  title = {Code Like Humans: A Multi-Agent Solution for Medical Coding},
  author = {Motzfeldt, Andreas Geert and Edin, Joakim and Christensen, Casper L. and Hardmeier, Christian and Maal{\o}e, Lars and Rogers, Anna},
  booktitle = {Findings of the Association for Computational Linguistics: EMNLP 2025},
  year = {2025},
  pages = {22612--22627},
  publisher = {Association for Computational Linguistics},
  doi = {10.18653/v1/2025.findings-emnlp.1231},
  url = {https://aclanthology.org/2025.findings-emnlp.1231/}
}

@misc{icd10cm2025,
  title = {{ICD-10-CM} Tabular List of Diseases and Injuries and Index to Diseases and Injuries, {FY} 2025},
  author = {{National Center for Health Statistics}},
  year = {2024},
  howpublished = {Centers for Disease Control and Prevention},
  url = {https://www.cdc.gov/nchs/icd/icd-10-cm/files.html}
}

@article{soroush2024poor,
  title = {Large Language Models Are Poor Medical Coders --- Benchmarking of Medical Code Querying},
  author = {Soroush, Ali and Glicksberg, Benjamin S. and Zimlichman, Eyal and Barash, Yiftach and Freeman, Robert and Charney, Alexander W. and Nadkarni, Girish N. and Klang, Eyal},
  journal = {NEJM AI},
  year = {2024},
  volume = {1},
  number = {5},
  doi = {10.1056/AIdbp2300040},
  url = {https://doi.org/10.1056/AIdbp2300040}
}

@inproceedings{lu-etal-2020-multi,
  title = {Multi-label Few/Zero-shot Learning with Knowledge Aggregated from Multiple Label Graphs},
  author = {Lu, Jueqing and Du, Lan and Liu, Ming and Dipnall, Joanna},
  booktitle = {Proceedings of the 2020 Conference on Empirical Methods in Natural Language Processing (EMNLP)},
  year = {2020},
  pages = {2935--2943},
  publisher = {Association for Computational Linguistics},
  doi = {10.18653/v1/2020.emnlp-main.235},
  url = {https://aclanthology.org/2020.emnlp-main.235/}
}

@inproceedings{yin-etal-2026-icdagent,
  title = {{ICDAGENT}: Empowering Agentic Large Language Models for Explainable Medical Coding},
  author = {Yin, Ziyi and Cao, Yuanpu and Wang, Ting and Chen, Jinghui and Ma, Fenglong},
  booktitle = {Proceedings of the 64th Annual Meeting of the Association for Computational Linguistics (Volume 1: Long Papers)},
  year = {2026},
  pages = {14131--14149},
  publisher = {Association for Computational Linguistics},
  doi = {10.18653/v1/2026.acl-long.643},
  url = {https://aclanthology.org/2026.acl-long.643/}
}

@article{ma2025mera,
  title = {Memorize and Rank: Elevating Large Language Models for Clinical Diagnosis Prediction},
  author = {Ma, Mingyu Derek and Wang, Xiaoxuan and Xiao, Yijia and Cuturrufo, Anthony and Nori, Vijay S. and Halperin, Eran and Wang, Wei},
  journal = {Proceedings of the AAAI Conference on Artificial Intelligence},
  year = {2025},
  volume = {39},
  number = {23},
  pages = {24786--24794},
  doi = {10.1609/aaai.v39i23.34660},
  url = {https://doi.org/10.1609/aaai.v39i23.34660}
}

\clearpage
\appendix
\section{Reproducibility Details}
\label{sec:appendix-reproducibility}

\subsection{Code Profile Construction}
\label{sec:appendix-code-profiles}

For each diagnosis code, we extract the chapter, block, category, and official
description. We then match candidate UMLS terms by normalized ICD code. The
process involves normalizing whitespace, discarding empty terms, eliminating
terms longer than 160 characters, and deduplicating terms in a case-insensitive
manner. When available, we select the exact match to the official description
as the preferred term. If there is no exact match, we choose the term with the
highest word-set overlap with the description, breaking ties by considering
length differences and lexical order. We also retain up to five additional
terms as synonyms. In cases where no terms are available, the preferred term
defaults to the official description.

We organize the profile into six lines in the following order: CHAPTER, BLOCK,
CATEGORY, DESCRIPTION, UMLS Terms, and SYNONYMS. Each line follows the
format \texttt{[FIELD] value}, with synonyms separated by semicolons.
Additionally, the FY2015 ICD-10-CM supplemental profiles may include curated
lexical paraphrases in the SYNONYMS field, recorded separately from UMLS terms.

For the ICD-10-CM mapping, we encode each six-field profile using
Qwen3-Embedding-4B.

Next, we construct Semantic IDs using hierarchical $K$-means clustering with
$K=128$ and $L=3$. At each node, the number of child clusters is determined by
taking the minimum of $K$, the group size, and the number of distinct embedding
vectors. Within each fitted partition, child-cluster indices are assigned by
sorting clusters according to the smallest ICD code they contain.

Once the three-level hierarchy is constructed, we append a UID token
only to paths that are shared by multiple codes. Codes with unique
three-token paths remain unchanged; however, codes that share paths receive
distinct UID tokens. This process ensures a one-to-one mapping from complete
SIDs to ICD codes.

\subsection{Prompt}
\label{sec:appendix-task-templates}

The following boxes present parameterized templates for SID and Raw tasks.
The field \texttt{\{code\_system\}} is instantiated as
\texttt{ICD-9} for MIMIC-III and \texttt{ICD-10-CM} for MIMIC-IV.
Other braced fields denote variable user inputs; model-specific
chat-template delimiters are omitted.
Each complete SID contains three path tokens and, only when needed to
disambiguate a shared path, a code-specific UID token. In the format below,
\texttt{[UID]} denotes this optional token; brackets are not generated.

\begin{promptbox}[skMain]{Note-to-SID}
You are a structured \texttt{\{code\_system\}} coding system.
\par\medskip
Given the discharge summary, output the complete set of supported diagnoses using only registered Semantic ID tokens.
\par\medskip
\textbf{Output format:}\par
\textless{}SID1\_...\textgreater{} \textless{}SID2\_...\textgreater{} \textless{}SID3\_...\textgreater{} \texttt{[UID]} \textless{}SEP\textgreater{} ... \textless{}SID1\_...\textgreater{} \textless{}SID2\_...\textgreater{} \textless{}SID3\_...\textgreater{} \texttt{[UID]} \textless{}END\textgreater{}
\par\medskip
\textbf{Rules:}\par
- Output Semantic ID tokens only.\par
- Each diagnosis must contain one SID1 token, one SID2 token, and one SID3 token in that order. Append the registered UID token when the codebook requires it, so that the complete SID identifies exactly one ICD code.\par
- Separate diagnoses with \textless{}SEP\textgreater{}.\par
- End the sequence with exactly one \textless{}END\textgreater{}.\par
- Do not output ICD codes, diagnosis names, explanations, headings, or any other text.\par
- Do not output anything after \textless{}END\textgreater{}.\par
\par\medskip
\prole{User input}\pvar{clinical\_document}
\end{promptbox}

\begin{promptbox}[skAux]{Profile-to-SID}
Convert the following disease profile into its Semantic ID. Output the three registered path tokens, followed by the code-specific UID token if required by the codebook, and then \textless{}END\textgreater{}, with no other text.
\par\medskip
\prole{User input}\pvar{code\_profile}
\end{promptbox}

\begin{promptbox}[skAux]{SID-to-Profile}
Describe the following Semantic ID using the six fields CHAPTER, BLOCK, CATEGORY, DESCRIPTION, UMLS Terms, and SYNONYMS. The complete Semantic ID identifies exactly one ICD code; output only that code's profile.
\par\medskip
\prole{User input}\pvar{sid}
\end{promptbox}

\begin{promptbox}[skMain]{Document-to-ICD (Raw)}
You are a \texttt{\{code\_system\}} diagnosis coding system.
\par\medskip
Given the discharge summary, output the complete set of \texttt{\{code\_system\}} diagnosis codes supported by the document.
\par\medskip
\textbf{Rules:}\par
- Use only information documented in the discharge summary.\par
- Include each code at most once.\par
- Separate codes with a comma followed by a space.\par
- Do not include code descriptions, explanations, headings, Markdown formatting, or text outside the block.\par
- If no diagnosis is supported, return \textless{}code\textgreater{}\textless{}/code\textgreater{}.
\par\medskip
\prole{User input}\pvar{clinical\_document}
\end{promptbox}

\subsection{Hyperparameter Tuning}
\label{sec:appendix-hyperparameters}

We tune the SID construction, auxiliary-task, and decoding hyperparameters using Qwen3-0.6B on the MIMIC-IV ICD-10-CM validation set. We consider the SID branching factor $K$, path depth $L$, auxiliary-task weight $\alpha$, and beam size $B$. The two auxiliary tasks receive equal weights, such that $(\lambda_1,\lambda_2,\lambda_3)=(1-2\alpha,\alpha,\alpha)$. We vary one hyperparameter at a time while fixing the others to $K=128$, $L=3$, $\alpha=0.1$, and $B=4$. We evaluate $K\in\{64,128,256\}$, $L\in\{2,3,4\}$, $\alpha\in\{0.05,0.1,0.2\}$, and $B\in\{1,2,4,8\}$. For beam-size tuning, we decode the same trained checkpoints with different values of $B$. All scores in this analysis are mean validation Micro-F1 over five random seeds.

\begin{figure}[t]
  \centering
  \includegraphics[width=\columnwidth]{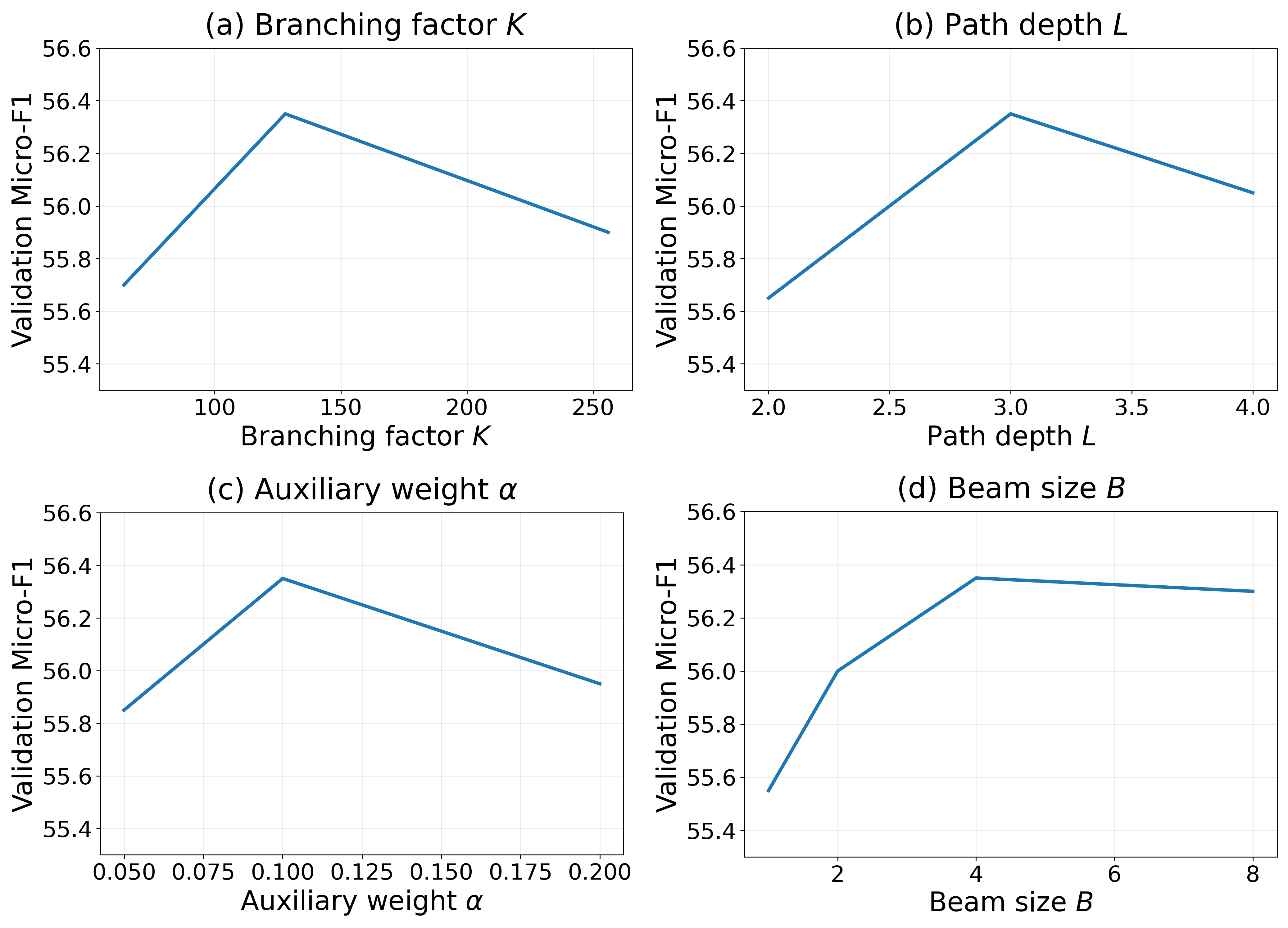}
  \caption{Hyperparameter tuning on the MIMIC-IV ICD-10-CM validation set using Qwen3-0.6B. Each panel varies one hyperparameter while fixing the others to $K=128$, $L=3$, $\alpha=0.1$, and $B=4$. Auxiliary-task weights are $(1-2\alpha,\alpha,\alpha)$. Values are mean Micro-F1 percentages over five random seeds.}
  \label{fig:hyperparameter-sensitivity}
\end{figure}

Figure~\ref{fig:hyperparameter-sensitivity} shows that $K=128$ and $L=3$ give the highest mean validation Micro-F1 within their respective search ranges. For the auxiliary-task weight, $\alpha=0.1$, corresponding to $(0.8,0.1,0.1)$, gives the highest score among the tested values. Performance also improves as the beam size increases from $1$ to $4$, with a small decrease at $B=8$. We therefore use $K=128$, $L=3$, $\alpha=0.1$, and $B=4$ in the main experiments.

\section{Structural Evaluation of SID Constructions}
\label{sec:appendix-sid-metrics}

We evaluate the native SID mappings produced by HKM, RK-Means,
and RQ-VAE on a common ICD-10-CM diagnosis catalog,
using the same input embeddings, $K=128$,
$L=3$, and construction seed 0. Each entry receives equal
weight. For RQ-VAE, we use native
nearest-neighbor assignments.
This excludes the RQ-VAE postprocessing step that reassigns final-level indices within collision groups while preserving earlier prefixes.

Let $\mathcal{C}$ denote the code catalog and $N=|\mathcal{C}|$. Each code
$c\in\mathcal{C}$ has a native SID path
$\mathbf z(c)=(z_1(c),\ldots,z_L(c))$ before optional UID disambiguation.
For $\ell=1,\ldots,L$, codes sharing the same prefix of length $\ell$
form a group. We denote the resulting partition by $\mathcal{G}_\ell$
and the group containing code $c$ by $g_\ell(c)$.

\paragraph{Adjusted Mutual Information.}
AMI measures the chance-adjusted agreement between first-level
SID groups and the ICD taxonomy. Let $y_{\mathrm{ch}}$ and
$y_{\mathrm{bl}}$ denote the chapter and block assignments of the catalog codes.
We report the average of the two AMI scores:
\begin{equation}
  \mathrm{AMI} =
  \frac{\operatorname{AMI}(\mathcal{G}_1,y_{\mathrm{ch}})
  + \operatorname{AMI}(\mathcal{G}_1,y_{\mathrm{bl}})}{2}.
\end{equation}
Each score uses arithmetic normalization.

\paragraph{Semantic Purity.}
Semantic Purity measures chapter consistency within first-level
SID groups. Let $n_{g,a}$ denote the number of codes in group
$g$ belonging to chapter $a$. We compute
\begin{equation}
  \mathrm{Purity} =
  \frac{1}{N}\sum_{g\in\mathcal{G}_1}\max_a n_{g,a}.
\end{equation}
Thus, the metric is the fraction of codes belonging to the
majority chapter of their group, with groups weighted by size.

\paragraph{Uniqueness.}
Uniqueness measures the fraction of codes uniquely identified
by their complete native SID paths before UID disambiguation. We define
\begin{equation}
  \mathrm{Uniqueness} =
  \frac{1}{N}\sum_{c\in\mathcal{C}}
  \mathbf{1}\!\left[|g_L(c)|=1\right].
\end{equation}
This counts codes in singleton full-path groups, rather than
the number of distinct paths.

\paragraph{Code Resolution Efficiency.}
Uniqueness considers only complete paths. CRE measures code-identity resolution across
prefix depths. We compute the entropy of the prefix partition at depth $\ell$ as
\begin{equation}
  H_\ell = -\sum_{g\in\mathcal{G}_\ell}
  \frac{|g|}{N}\log\frac{|g|}{N},
\end{equation}
and define
\begin{equation}
  \mathrm{CRE} = \frac{1}{L}\sum_{\ell=1}^{L}
  \frac{H_\ell}{\log N}.
\end{equation}
Equivalently,
\begin{equation}
  H_\ell = \log N - \frac{1}{N}\sum_{c\in\mathcal{C}}
  \log |g_\ell(c)|.
\end{equation}
Under a uniform prior over catalog codes, $H_\ell/\log N$ is the fraction
of code-identity uncertainty resolved by a length-$\ell$ prefix.
CRE averages this fraction across depths. Unlike Uniqueness, it also
captures partial resolution when a prefix narrows the candidate group
without uniquely identifying a code.
When complete paths are equally discriminative, resolving code identity
at earlier depths yields a higher CRE.

\end{document}